\documentclass[letterpaper, 10 pt, conference]{ieeeconf}  

\IEEEoverridecommandlockouts                              

\usepackage{graphics} 
\usepackage{epsfig} 
\usepackage{marvosym}
\usepackage{makecell}
\usepackage{multirow}
\usepackage{multicol}
\usepackage{hyperref}

\title{\LARGE \bf
Enable Natural Tactile Interaction for Robot Dog based on Large-format Distributed Flexible Pressure Sensors
}

\author{Lishuang Zhan$^{1}$, Yancheng Cao$^{1}$, Qitai Chen$^{1}$, Haole Guo$^{1}$,\\ Jiasi Gao$^{1}$, Yiyue Luo$^{2}$, Shihui Guo$^{3}$, Guyue Zhou$^{1}$ and Jiangtao Gong$^{1}$\textsuperscript{\Letter}
\thanks{$^{1}$Institute for AI Industry Research (AIR), Tsinghua University, 10080, Haidian District, Beijing, P.R.China.
        {\tt\small lastnamefirstname@air.tsinghua.edu.cn}}
\thanks{$^{2}$Computer Science and Artificial Intelligence Laboratory (CSAIL), Massachusetts Institute of Technology.}
\thanks{$^{3}$School of Informatics, Xiamen University.}
\thanks{Supplementary at: \href{https://github.com/AIR-DISCOVER/Human-Robot-Dog-Tactile-Interaction}{https://github.com/AIR-DISCOVER/Human-Robot-Dog-Tactile-Interaction}}
}

\begin{document}

\maketitle
\thispagestyle{empty}
\pagestyle{empty}

\begin{abstract}
Touch is an important channel for human-robot interaction, while it is challenging for robots to recognize human touch accurately and make appropriate responses.
In this paper, we design and implement a set of large-format distributed flexible pressure sensors on a robot dog to enable natural human-robot tactile interaction. 
Through a heuristic study, we sorted out 81 tactile gestures commonly used when humans interact with real dogs and 44 dog reactions. 
A gesture classification algorithm based on ResNet is proposed to recognize these 81 human gestures, and the classification accuracy reaches 98.7\%. 
In addition, an action prediction algorithm based on Transformer is proposed to predict dog actions from human gestures, reaching a 1-gram BLEU score of 0.87. 
Finally, we compare the tactile interaction with the voice interaction during a freedom human-robot-dog interactive playing study. 
The results show that tactile interaction plays a more significant role in alleviating user anxiety, stimulating user excitement and improving the acceptability of robot dogs. 
\end{abstract}

\section{INTRODUCTION}
Touch is one of the most primitive sensory channels of organisms. 
Harlow’s rhesus monkey experiment~\cite{harlow1958nature} shows that comfortable physical contact plays an important role in natural emotion development. 
Tactile interaction is also of great significance for the relationship between humans and pet robots. 
Studies have found that affective connection with pet dogs can reduce humans’ loneliness~\cite{rew2000friends}, and tactile contact with them can even benefit humans’ health~\cite{vormbrock1988cardiovascular}. 
Since the development of pet robots, they have been able to show some animal-like behaviors when interacting with humans, including affective expressions~\cite{lee2016designing}. 
In particular, tactile interaction will affect human emotions, which will be reflected in their different gestures when touching robot dogs~\cite{yohanan2012role}. 

However, compared with the popular visual and auditory technologies, the tactile-based human-robot interaction technology is relatively immature.
Previous studies in the field of robot tactile sensing mainly focus on local perception, such as mechanical arm~\cite{massari2022functional, sarwar2021large}, gripper~\cite{zlokapa2022integrated, li2022tata}, local robot skin~\cite{eom2021embedded}, etc.; and the whole-body tactile sensing systems are primarily based on discrete~\cite{shibata2001mental} or low-density~\cite{chang2010gesture, flagg2013affective} sensor layouts.
Up to now, it has been challenging and seldom implemented to enable large-format and high-density tactile sensing for robots. 
Moreover, there are few studies on the human-robot tactile interaction pipeline, \textit{i.e.}, from human touch to robot feedback.

To fill these gaps, we design and implement a real-time interactive system based on large-format distributed flexible pressure sensors in this paper. 
By sticking a layer of piezoresistive film sensors on the skin of a robot dog, the sensors cover the robot’s skin in a large area and a special shape, including the head, back, hips, legs and other parts. 

Based on existing work collecting the gesture set when humans interact with tactile creatures~\cite{yohanan2012role,flagg2013affective,wang2021organization}, we carried out a heuristic study, including questionnaire distribution and video collection, to examine humans’ gesture preference in tactile interaction with dogs and dogs’ feedback actions. 
Through questionnaire screening and video analysis, we sorted out 13 human gestures and 11 dog body parts, and combined them to obtain 81 tactile gestures. 
In addition, we summarized 44 dog actions from the videos. 

We use the gesture recognition algorithm based on ResNet~\cite{he2016deep} to classify the above 81 human gestures, and the accuracy reaches 98.7\%. 
These gestures will trigger different dog actions, which is realized by Transformer~\cite{vaswani2017attention}, reaching a 1-gram BLEU~\cite{papineni2002bleu} score of 0.87.

\begin{figure}[t]
\centering
  \includegraphics[width=.485\textwidth]{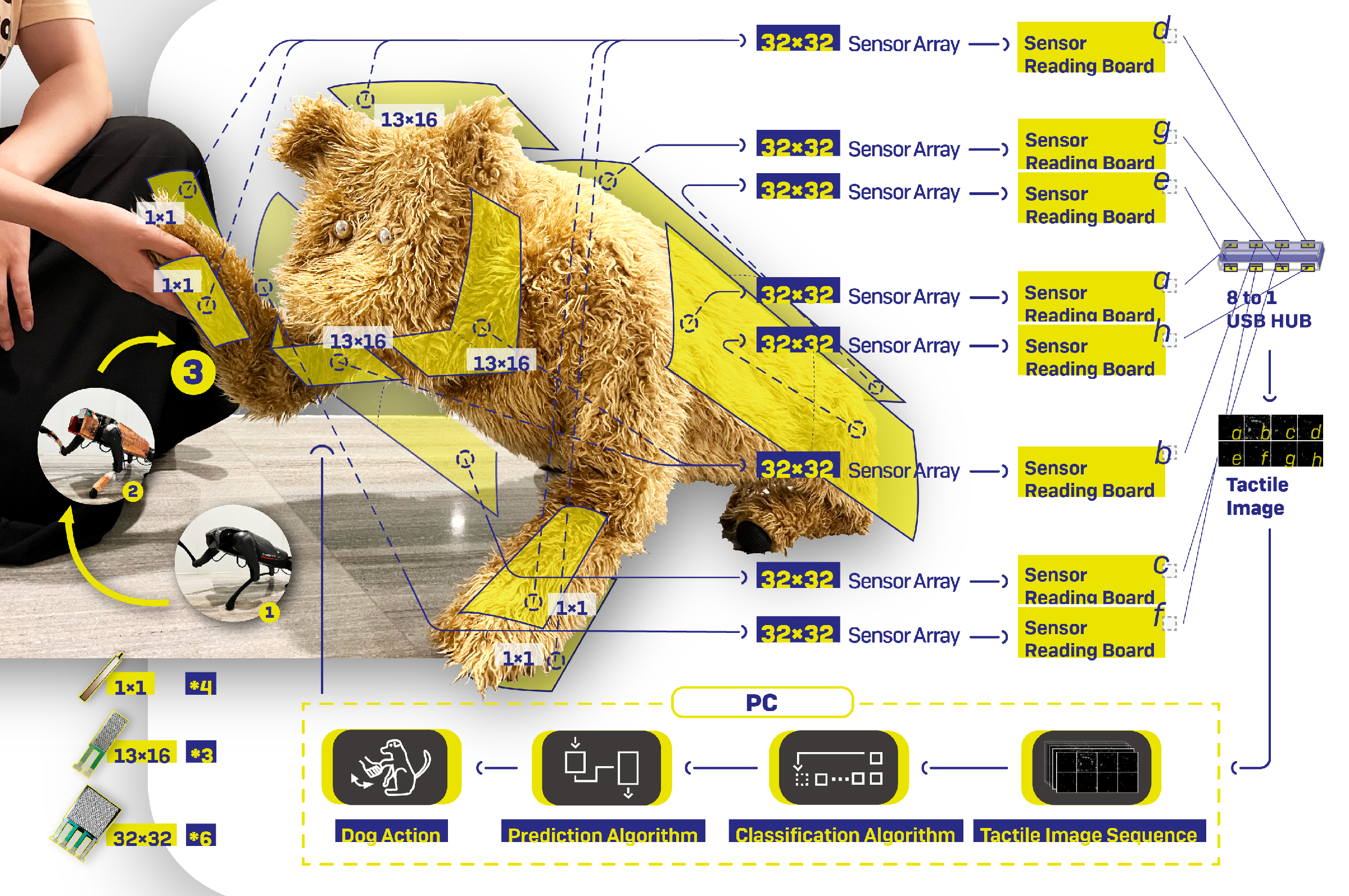}
  \caption{The overview of hardware framework and algorithm pipeline.}
  \label{fig:mainImage}
\end{figure}

Finally, we conducted a compared experiment to verify the superiority of our tactile interaction over the voice interaction on human emotion regulation. 
The results show that compared with the voice interaction, the participants in the tactile group have more significant emotional progress after the experiment and are willing to spend more time playing with the robot dog.

Therefore, the contributions of this paper are as follows:
\begin{itemize}
\item A real-time interactive system based on large-format distributed flexible pressure sensors, enabling human-robot-dog tactile interaction from human touch to dog reactions.
\item Eighty-one human tactile gestures commonly used when interacting with dogs and 44 dog feedback actions are summarized through a heuristic study.
\item A novel algorithm architecture is proposed, with a human gesture classification algorithm based on ResNet and a dog action prediction algorithm based on Transformer, which achieves a classification accuracy of 98.7\% and a 1-gram BLEU score of 0.87, respectively.
\item A compared experiment that reveals human-robot-dog tactile interaction is significantly more effective in promoting users’ positive emotions than voice interaction.
\end{itemize}

\section{Hardware}
The hardware system consists of the robot dog, the piezoresistive film sensors and the sensor reading boards. They are described in detail below.

\subsection{Piezoresistive Film Sensor}
The piezoresistive film sensors we use are off-shelf provided by \textit{Roxifsr}\footnote{https://world.taobao.com/item/609970762251.htm}. 
It is made by transferring the force-sensitive materials, silver paste and other materials to the flexible film substrate through the precision printing process, and then drying and curing. 
When pressure is applied to the sensor, the resistance decreases as the pressure increases. 

As shown in Fig.\ref{fig:mainImage}, a total of 13 pressure sensors with three sizes are applied to the robot dog: i) six 32 $\times$ 32 sensors are attached to the body; ii) three 13 $\times$ 16 sensors attached to the head, the cheeks and the chin; iii) four single-point sensors attached to the forelimbs.

The 32 $\times$ 32 sensor arrays provide 1024 sensing points within 16 $\times$ 16 cm$^2$ area, with a maximum measurement range of 2$kg$. 
The 13 $\times$ 16 sensor arrays are cut out from the 32 $\times$ 32 ones, all of which share the same parameters. 
The single-point pressure sensors are located at the front and back of each forelimb, with a maximum measurement range of 25$kg$. 
Besides, the top and bottom of all sensors are wrapped with aluminum foil and connected with copper foil tape, forming a shield to reduce noise. 

\subsection{Sensor Reading Board}
There are eight sensor reading boards installed in the robot dog, six for the body sensors, one for the cheeks and chin sensors, and one for the head and forelimbs sensors. 
The sensor reading board adopts STM32F072R8T6 as MCU (Micro Control Unit), 8x CD4051R as multiplexer to select rows and columns, and LM321M5 as voltage amplifier. 
All sensor reading boards are wired to the PC and powered by external power through an 8-in-1 USB hub. 
They collect data at a frame rate of 10$Hz$, and the serial ports realize data synchronization through broadcast communication.

\subsection{Robot Dog}
The robot dog we use is the \textit{CyberDog} provided by \textit{MI}\footnote{https://www.mi.com/cyberdog}. 
GRPC communication is adopted between the PC and \textit{CyberDog}, which is guaranteed to be in a LAN through a wired connection. 
Tactile input is calculated as action instruction on PC, and transmitted to \textit{CyberDog} for execution. 
The action arrangement of \textit{CyberDog} is based on ROS2, and different designs are realized by modifying the action parameters of \textit{.toml} files. 
Twelve brushless motors in the robot dog are controlled by calling the \textit{.toml} file, changing the postures and realizing the action control of the robot dog. 
In addition, for the beautiful appearance and comfortable touch, we customized a fur cover for the robot dog.


\section{Heuristic Study}

\begin{figure}[ht]
\centering
  \includegraphics[width=.5\textwidth]{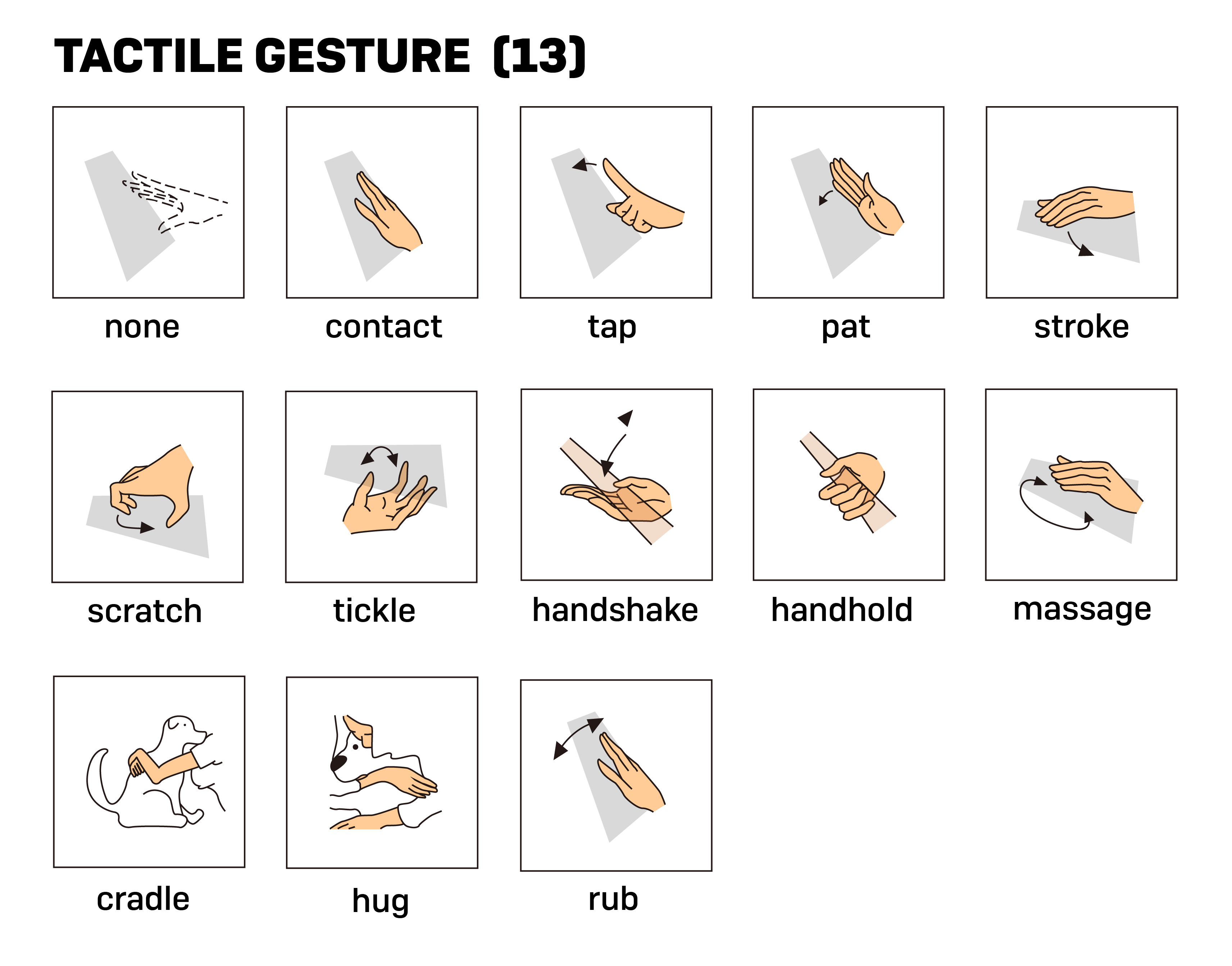}
  \caption{Thirteen human tactile gestures.}
  \label{fig:gesturesImage}
\end{figure}

\begin{figure}[ht]
\centering
  \includegraphics[width=.45\textwidth]{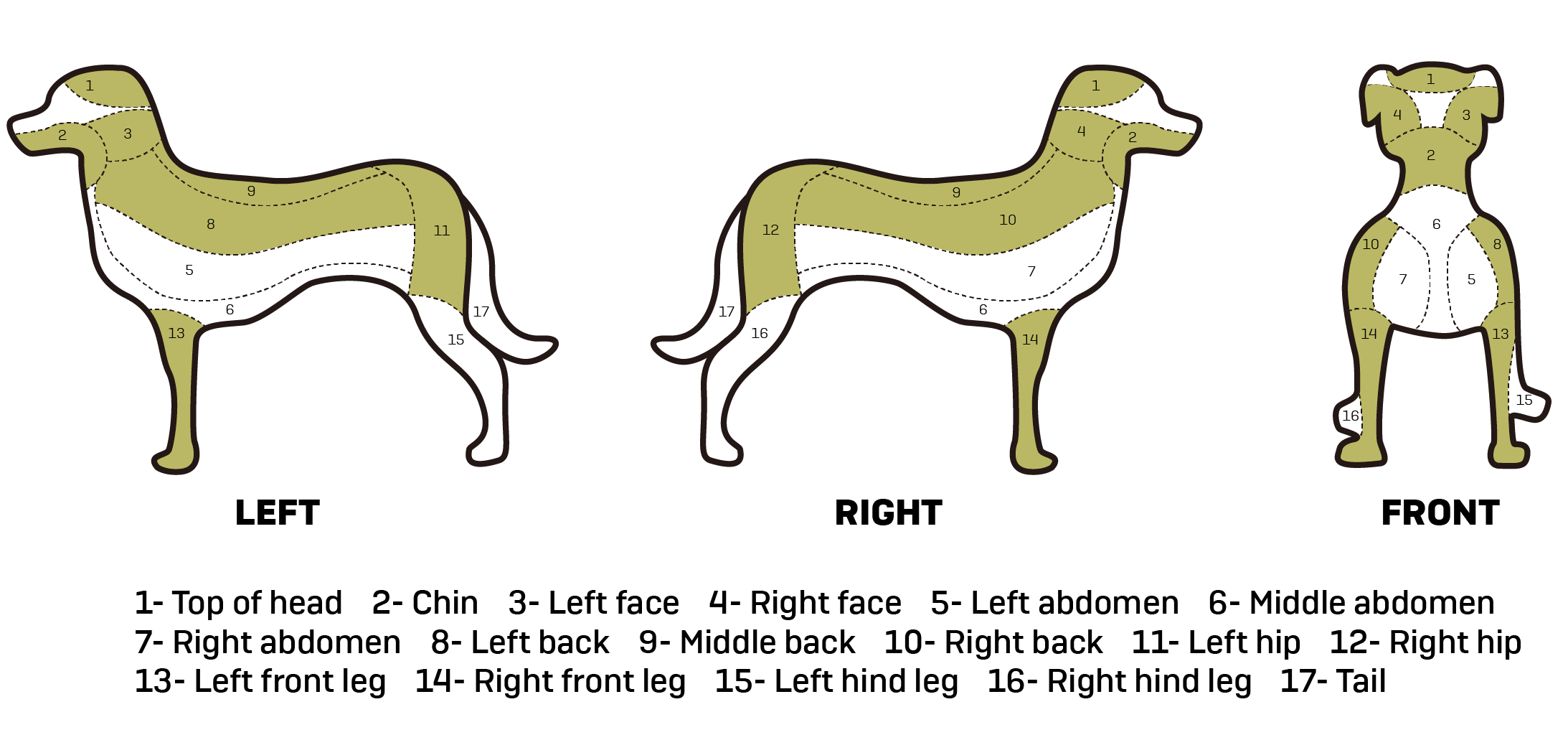}
  \caption{Eleven tactile interaction parts on the dog (marked with yellow).}
  \label{fig:touchingParts}
\end{figure}

\begin{figure*}[t]
\centering
  \includegraphics[width=1.0\textwidth]{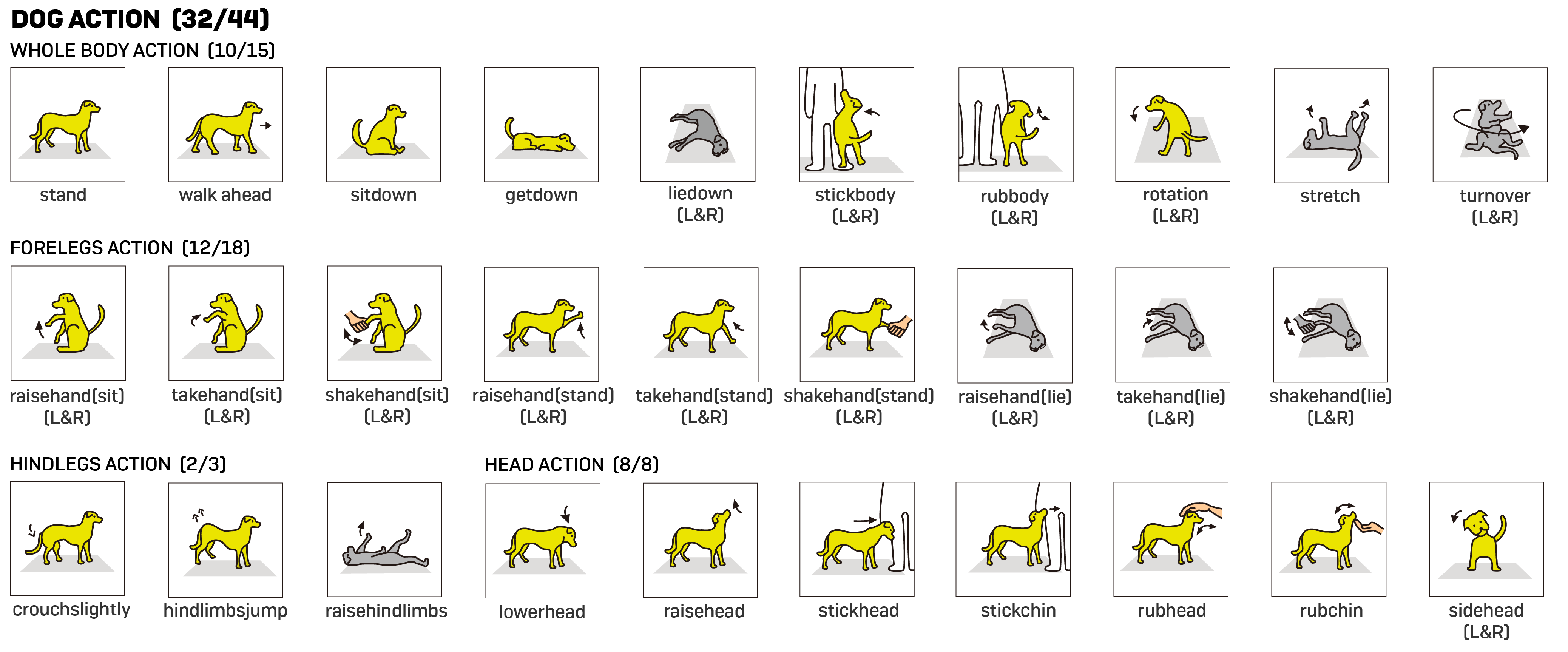}
  \caption{Forty-four dog actions in four categories, of which 32 actions can be completed by the robot dog at present (marked with yellow).}
  \label{fig:actionsImage}
\end{figure*}

In this paper, we define tactile gestures as the combination of human gestures and interaction parts on the dog, such as stroking the head. 
These tactile gestures are dynamic and usually contain human feelings or intentions when executed. 
In order to explore the gesture preference of dog owners for tactile interaction with their pet dogs in real life and dogs’ different reactions to different gestures, we designed a questionnaire and collected tactile interaction videos between dog owners and their dogs for research.

\subsection{Participants}
This experiment is aimed at people who have experience keeping dogs (who have had dogs or are keeping dogs). 
According to the daily interaction between these people and their pet dogs, we try to find the tactile interaction pattern between humans and robot dogs. 
In addition to collecting questionnaires, we contacted the users who left their contact information and expressed their willingness to participate in further experiments. 
They were asked to submit one or more videos of touching dogs in their daily lives, with a total duration of about 10 minutes. 
Finally, 109 questionnaires passed the attention test and were retained. 
Among them, 26 users submitted their videos, which added up to 208 minutes.

\subsection{Task and Procedure}
According to the collected 109 questionnaires, we sorted out users’ preferred gestures and obtained a group of tactile gestures. 
In addition, by analyzing the 26 interaction videos, we supplemented some common gestures and extracted dog feedback actions under corresponding triggers.

\subsubsection{Human Tactile Gestures}
Referring to the dictionary of 30 human gestures of touching tactile creatures compiled by Yohanan and MacLean \cite{yohanan2012role}, we summarized the tactile gesture set of human-robot-dog interaction in this paper through the following screening scheme.

\begin{itemize}
\item First of all, we screened out four of the 30 gestures that could not be done to the robot dog because of its heavy weight: \textit{Hold}, \textit{Lift}, \textit{Swing} and \textit{Toss}.
\item In the questionnaire, we invited users to rate these 26 tactile gestures using the five-point Likert scale. The scores from 1 to 5 represent the increasing preference. After summary and analysis, we selected 15 gestures that all users scored more than 3 points on average.
\item Gestures that can not be detected by the piezoresistive sensors and are potentially harmful to the posture balance of the robot dog were also deleted, that is, \textit{Pick}, \textit{Pinch}, \textit{Finger Idly}, \textit{Tremble} and \textit{Rock}.
\item \textit{Handhold} and \textit{Handshake} were added to the gesture set because they frequently appeared in videos.
\item The gesture ``\textit{none}'' was added to describe the situation where no tactile interaction occurs. 
\end{itemize}

Finally, a total of 13 gestures (see Fig.\ref{fig:gesturesImage}) were included in the tactile gesture set. 
By observing the videos, we found that the dog body parts touched by users could be roughly divided into five categories: head, back, abdomen, hips and forelimbs. 
However, the interaction on the abdomen usually occurs when the dog lies on his back, and we have not arranged any related actions (\textit{i.e.}, the actions marked with gray in Fig.~\ref{fig:actionsImage}) at present because of the raised back. 
Therefore, for the current implementation, abdominal interaction was considered to be excluded, and the interaction parts were divided into 11 (see Fig.~\ref{fig:touchingParts}) after detailed classification. 
Finally, 81 human gestures towards the robot dog were compiled by combining 13 tactile gestures with 11 interaction parts.

\subsubsection{Dog Feedback Actions}
We collected the dog action set by observing and analyzing the real reactions of dogs to human touch in the videos. The followings are our processes and rules for screening and summarizing dog actions.

\begin{itemize}
    \item Sort out the common actions repeated in different dogs. According to the changes in the position, direction and stretching degree of specific body parts of the dogs, we summarized actions like standing, sitting and lying down, etc. Besides, we concluded actions such as sticking and rubbing according to the touch strength and angle, and whether there was relative sliding.
    \item Actions that have only appeared a few times but are interesting were also kept in our list, such as crouching slightly and hindlimbs jumping.
    \item Screen out actions that the robot dog cannot perform or are dangerous, such as twisting the neck or pouncing on someone.
    \item Some slight actions, such as limb shaking, were not considered.
    \item Eliminate interactions triggered by senses other than touch, such as hearing or vision.
    \item Screen out interactions caused by other items, not the human body.
    \item According to the body parts and the action fineness, dog actions were divided into four categories: whole body actions, forelimbs actions, hindlimbs actions and head actions.
\end{itemize}

Finally, we came to the conclusion that dog feedback actions to human tactile gestures could be divided into four categories, with a total of 44 actions (see Fig.\ref{fig:actionsImage}).

\subsection{Discussion}
The interaction between humans and dogs reflected in the videos is quite rich. 
Human gestures and dog actions usually show a reasonable correspondence, but they are not completely consistent. 
For example, when a person strokes a dog’s head, the dog may bow its head or rub its head against the person’s hand. 
These reasonable but diverse reactions are a great pleasure for humans to play with their pet dogs, which provides us with a powerful inspiration to realize the tactile feedback mechanism.

In short, based on the questionnaires and videos collected, we summarized 13 human gestures and 11 interaction parts on the dog, making a total of 81 tactile gestures. With regard to dog feedback actions to human tactile gestures, we analyzed their performance in the videos and finally compiled an action set containing 44 common actions.



\section{Algorithm Architecture}

\begin{figure*}[t]
\centering
  \includegraphics[width=1.0\textwidth]{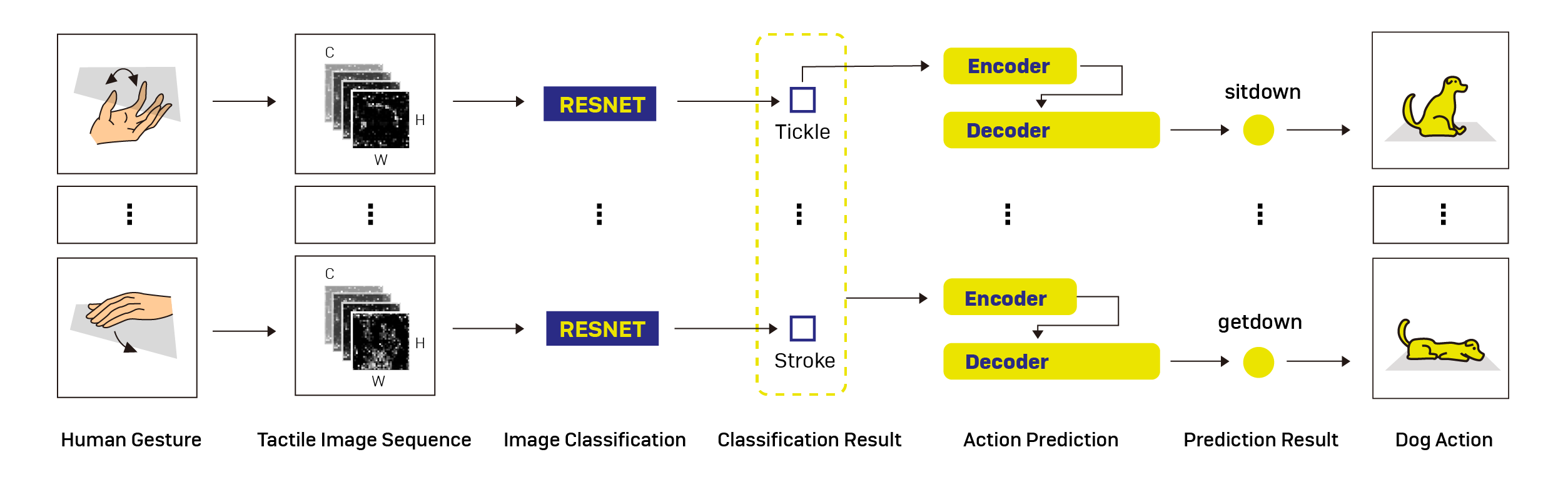}
  \caption{The algorithm architecture with a tactile gesture classification model based on ResNet and a dog action prediction model based on Transformer.}
  \label{fig:OverallStructure}
\end{figure*}

The overall algorithm architecture from human tactile gestures to dog feedback actions is shown in Fig.~\ref{fig:OverallStructure}, which consists of two main steps: human tactile gesture classification and dog feedback action prediction. 
Tactile gesture recognition is implemented using ResNet~\cite{he2016deep} architecture, with tactile image sequence as input and tactile gesture classification result as output. 
Dog action prediction is implemented using Transformer~\cite{vaswani2017attention} architecture, which is actually regarded as a classic translation task. The prediction algorithm, Transformer, is used to ``translate'' human gesture ``language'' into dog action ``language'', taking human gesture ``word (s)'' (sequence) as input and dog action ``word (s)'' (sequence) as output.

\subsection{Tactile Gesture Recognition}
\subsubsection{Dataset}
By combining 13 human gestures with 11 dog body parts, we get 81 categories of tactile gestures.
The gesture dataset includes 14,580 image sequences, including 180 in each category. For the dataset, 90$\%$ is kept for training while the remaining is for testing.
In the data collection process, each gesture was repeated 20 times by each person, with 9 people participating, {\it i.e.}, 81 categories $\times$ 20 times $\times$ 9 people = 14,580 image sequences. 
Each image sequence includes 20 frames of tactile images, that is, a total of 291,600 tactile images. 
The resolution of each tactile image is 64 $\times$ 128, composed of eight 32 $\times$ 32 sensor data.

\subsubsection{Classification Algorithm}
We formulate our gesture recognition task as a multiclass ({\it i.e.}, 81 classes) classification problem. 
Accordingly, we build a variant of the ResNet34 deep residual CNN by adjusting its input layer to match our input image size of H $\times$ W $\times$ C (H $=$ 64, W $=$ 128, C $=$ 20) and its output layer to match our output dimension of 81. Then, we can train our ResNet34 model in a supervised manner using our dataset. We implement this model in Python using the {\it pytorch} framework.

\subsubsection{Results}
Our gesture classification model achieves 98.7\% accuracy on the 81-class classification task on the test set.

\subsection{Dog Action Prediction}
\subsubsection{Vocabulary Construction}
We formulate our action prediction task as a translation problem. Therefore, we have constructed two vocabularies, one for human gesture names and the other for dog action names. The former contains 81 human gestures, while the latter contains 40 dog actions.

\subsubsection{Dataset}
We collected a total of 208-minute videos from 26 participants. 
These videos showed people’s tactile gestures when they interacted with dogs and dogs’ feedback actions. 
We recorded the human gestures and the dog reactions respectively, and constructed a dataset of 1212 human gestures-dog actions interaction sequences. 
Five-sixths of the dataset is kept for training, while the remaining is for testing.

\subsubsection{Prediction Algorithm}
We adopt the Transformer model, which relies on the multi-head self-attention mechanism to realize the translation task of sequence to sequence. 
The input of the network is a sequence of human gesture words with a maximum length of 20 (including a beginning word and an ending word), representing the current and the previous human gestures to consider the context. 
The output of the network is the corresponding sequence of dog action words. We take the last word of the output sequence as the prediction result of the dog action.

\subsubsection{Results}
BLEU~\cite{papineni2002bleu} is used to evaluate the performance of the action prediction algorithm. On the translation task from human gestures to dog actions, our prediction model achieved a 1-gram BLEU score of 0.87.

\section{Compared Experiment}
In order to evaluate the impact of tactile interaction with the robot dog on humans’ emotions and attitudes towards robot dogs, we carried out an experiment to compare our tactile interaction scheme with the voice scheme.

\subsection{Participants}
Sixteen participants (Female $=$ 7, avg. age $=$ 26.2 years) were invited to participate in our experiment and randomly divided into two groups, each group with 8 participants: 

1) Group A (experimental group): Participants interact with the robot dog by touching it;

2) Group B (control group): Participants interact with the robot dog using voice instruction, \textit{e.g.}, ``sit down''. The voice interaction is based on the Wizard of Oz method~\cite{green1985rapid};

Our research was IRB approved by the local institution of Tsinghua University.
All participants signed the informed consent form before the experiment, and they were paid accordingly after the experiment.

\subsection{Scales Design}
We designed three scales to evaluate the participants’ emotions and attitudes toward robot dogs before and after the test.

The first one is an attitude scale of humans towards robot dogs, merged with the questions which are suitable for robot dogs of Companion Animal Bonding Scale~\cite{poresky1987companion}, Pet Attitude Scale~\cite{templer1981construction,canfield2004modification} and Animal Attitude Scale~\cite{herzog2015brief}. It is scored by the five-point Likert scale, and the participants’ acceptance of robot dogs is positively correlated with the average score of all questions.

In order to evaluate the changes in participants’ depression and anxiety, we design the second scale, consisting of the Self-Rating Depression Scale~\cite{zung1965self} and the Self-Rating Anxiety Scale~\cite{zung1971rating}. The title of this scale is visible to participants as the Emotion Self-rating Scale, rated by the four-point Likert scale. The higher the average score of all questions, the stronger the participants’ depression and anxiety.

The third scale comes from Russell’s theory, which holds that emotion comprises two bipolar and orthogonal dimensions~\cite{russell1980circumplex, russell1989affect}. One dimension, valence, is described by the continuum from unpleasant to pleasant, and the other dimension, arousal, ranges from deactivated to activated. These two dimensions are represented by the horizontal axis and vertical axis respectively, and the positive direction of the coordinate axis indicates more positive valence and higher arousal level, with a maximum coordinate value of 3.

Finally, the four indicators indicated by these three scales: humans’ attitude towards robot dogs, humans’ depression and anxiety intensity, valence level and arousal level are used to evaluate the changes in participants’ pre-and post-test status.

\subsection{Procedure}
The experiment consists of three parts: pre-test, formal test and post-test. 
In the pre-and post-test, participants filled out the above three scales, respectively. In the formal test, participants were allowed to interact freely with the robot dog. The maximum interaction time was set to 20 minutes, and the interaction between participants and the robot dog would not be interrupted within 20 minutes until they took the initiative to stop. 
The two groups of experiments were conducted separately, and the participants in each group participated in the experiments in turn without communication.

\subsection{Results}
Before significance tests, the experimental data are tested for normality. The data conforming to the normal distribution is tested by T-test; otherwise, the Mann-Whitney U-Test.
The null hypotheses include i) there is no difference in user indicators in the pre-and post-test; ii) there is no difference in user indicators between group A and group B.

\subsubsection{Humans’ Attitude towards Robot Dogs}

It can be seen from Table.~\ref{table1} that participants’ attitudes towards the robot dog have changed significantly in the pre-and post-test of the experimental group. However, the attitudes of participants in the control group hardly changed. From the significance test results between groups, it can also be seen that the attitudes of the two groups of participants are roughly the same in the pre-test, but there are significant differences in the post-test. 
As P4 in the experimental group said, “\textit{I thought it was a machine at first sight, but after playing for a while, I thought he was like a pet dog.}" However, P7 of the control group indicated that he felt he was playing a game instead of playing with a pet.

\begin{table}[h]
    \caption{Significance test of the attitude}
    \label{table1}
    \centering
    \begin{tabular}{|c||c||c||c||c|}
    \hline
        ~ & Group A & Group B & Value & \textit{P} \\ \hline
        pre-test & 3.41$\pm$0.29 & 3.27$\pm$0.31 & t=0.904 & \textit{P}=0.191 \\ \hline
        post-test & 3.88$\pm$0.32 & 2.99$\pm$0.53 & t=4.030 & \textit{P}\textless0.001*** \\ \hline
        Value & t=-3.067 & t=1.292 & ~ & ~ \\ \hline
        \textit{P} & \textit{P}=0.004** & \textit{P}=0.109 & ~ & ~ \\ \hline
    \end{tabular}
\end{table}

\subsubsection{Humans’ Depression and Anxiety Intensity}

The results in Table~\ref{table2} show that the depression and anxiety of the participants in the experimental group were slightly relieved after interacting in touch with the robot dog, but no change occurred in the control group. 
The slight change may be due to the continuity of depression and anxiety that is difficult to change significantly in a short time. 
Nevertheless, there were significant differences in participants’ emotions between the experimental group and the control group in the post-test while without any difference in the pre-test; that is, participants’ depression and anxiety decreased even more after tactile interaction with the robot dog. 
Many participants in the tactile group mentioned that they felt healed.

\begin{table}[h]
    \caption{Significance test of the depression and anxiety}
    \label{table2}
    \centering
    \begin{tabular}{|c||c||c||c||c|}
    \hline
        ~ & Group A & Group B & Value & \textit{P} \\ \hline
        pre-test & 1.90$\pm$0.46 & 1.91$\pm$0.36 & t=-0.030 & \textit{P}=0.488 \\ \hline
        post-test & 1.61$\pm$0.22 & 1.88$\pm$0.31 & t=-1.937 & \textit{P}=0.037* \\ \hline
        Value & t=1.604 & t=0.184 & ~ & ~ \\ \hline
        \textit{P} & \textit{P}=0.065 & \textit{P}=0.428 & ~ & ~ \\ \hline
    \end{tabular}
\end{table}

\subsubsection{Valence Level}
Tactile interaction with the robot dog can significantly make participants feel pleasant, and voice interaction can also play a slight role (see Table~\ref{table3}). 
Voice interaction with the robot dog gave participants the experience of playing games, which made them feel good. 
However, the intimacy and surprise accompanying tactile interaction could quickly make participants’ emotions reach a higher point. 
P5 of the experiment group gave a ``cool'' exclamation all the time during the experiment.

\begin{table}[h]
    \caption{Significance test of the valence}
    \label{table3}
    \centering
    \begin{tabular}{|c||c||c||c||c|}
    \hline
        ~ & Group A & Group B & Value & \textit{P} \\ \hline
        pre-test & 0.60$\pm$0.93 & 1.13$\pm$0.66 & t=-1.306 & \textit{P}=0.106 \\ \hline
        post-test & 1.73$\pm$0.40 & 1.63$\pm$0.81 & \textit{U}=30.000 & \textit{P}=0.878 \\ \hline
        Value & \textit{U}=58.500 & t=-1.358 & ~ & ~ \\ \hline
        \textit{P} & \textit{P}=0.003** & \textit{P}=0.098 & ~ & ~ \\ \hline
    \end{tabular}
\end{table}

\subsubsection{Arousal Level}
After tactile interaction with the robot dog, the arousal level of the participants was significantly activated. However, voice interaction has no significant impact on participants’ arousal level (see Table~\ref{table4}). 
In addition, there was an interesting finding that the arousal level of participants who gave positive comments was not always improved. Some participants mentioned that the intimate and comfortable tactile interaction makes them feel at ease and become more relaxed, accompanied by a decrease in the arousal level.

\begin{table}[h]
    \caption{Significance test of the arousal}
    \label{table4}
    \centering
    \begin{tabular}{|c||c||c||c||c|}
    \hline
        ~ & Group A & Group B & Value & \textit{P} \\ \hline
        pre-test & -0.16$\pm$1.24 & 0.04$\pm$0.67 & t=-0.401 &\textit{P}=0.347 \\ \hline
        post-test & 1.05$\pm$1.70 & 0.46$\pm$1.31 & \textit{U}=22.000 & \textit{P}=0.328 \\ \hline
        Value & \textit{U}=51.000 & t=-0.815 & ~ & ~ \\ \hline
        \textit{P} & \textit{P}=0.050* & \textit{P}=0.214 & ~ & ~ \\ \hline
    \end{tabular}
\end{table}

\subsection{Discussion}
Among the above four indexes, the tactile interaction scheme has achieved significantly better results than the voice scheme. 
In addition, we recorded the experimental time of the participants. The average interaction time of the experimental group was 12.5 minutes, while that of the control group was 5.5 minutes. 
Most participants in the control group would stop the experiment after executing all the voice instructions once. On the contrary, the action feedback from natural tactile interaction could keep participants fresh. 
In a word, the results of the compared experiment strongly prove that the immersion experience brought by touch is of great significance to natural human-robot interaction.

\section{CONCLUSION and Future Work}
\subsection{Natural Tactile Interaction for Robot Dog}
In this paper, we implement a human-robot-dog tactile interaction system based on large-format distributed flexible pressure sensors.
Through a heuristic research, we sort out 81 human tactile gestures and 44 dog feedback actions. 
A novel two-step algorithm architecture is proposed: a gesture classification algorithm based on ResNet to recognize these 81 human gestures with a classification accuracy of 98.7\%, and an action prediction algorithm based on Transformer to predict dog actions reaching a 1-gram BLEU score of 0.87. 
Finally, through a compared experiment, it is proved that compared with voice interaction, natural tactile interaction plays a more significant role in relieving user anxiety, stimulating user excitement and improving the acceptability of robot dogs.
Supplementary materials on gesture definitions and custom scales are available at: \href{https://github.com/AIR-DISCOVER/Human-Robot-Dog-Tactile-Interaction}{https://github.com/AIR-DISCOVER/Human-Robot-Dog-Tactile-Interaction}.

\subsection{Future Work}
\subsubsection{Richer Dog Actions and Whole-Body Tactile Sensing}
The robot dog’s back was padded to get more space to accommodate the sensor reading boards and wires embedded in its body. 
As a result, it could not perform back-pressing actions such as tumbling, and no sensor was attached to the abdomen. 
In addition, the actions involving a flexible neck or tail were not considered due to the dog’s rigid shape. 
In the future, we would explore a better hardware assembly scheme and realize richer dog actions as well as the whole-body tactile sensing system on the robot dog.

\subsubsection{Multi-modal Interaction}
Multi-modal interaction scheme is of great significance in improving the naturalness and efficiency of human-robot interaction. 
Some participants mentioned it would be great if the robot dog could understand what they said while understanding tactile gestures.
Therefore, we would expand voice and vision channels to further improve the system functionality and flexibility.

\subsubsection{End-to-end Model Architecture}
In the current implementation, a two-step model architecture, that is, gesture recognition and action prediction, is used for the human-robot-dog tactile interaction pipeline to make it more interpretable. 
An end-to-end model may be a meaningful and challenging attempt in further work.

\addtolength{\textheight}{-9cm}  

\bibliographystyle{IEEEtran}
\bibliography{IEEEfull,root}




\end{document}